\documentclass[10pt,twocolumn,letterpaper]{article}

\usepackage{cvpr}              % To produce the CAMERA-READY version
\usepackage{multirow}

\usepackage[T1]{fontenc}
\usepackage{inconsolata}
\usepackage{enumitem}
\usepackage[most]{tcolorbox}
\usepackage{xcolor}
\definecolor{promptbg}{RGB}{246,248,251}

\tcbset{promptstyle/.style={
  enhanced,
  width=\linewidth,
  colback=promptbg,
  frame hidden,
  arc=2pt,
  left=7pt, right=7pt, top=6pt, bottom=6pt,
  fontupper=\footnotesize\ttfamily\raggedright,
  before upper={\setlength{\parskip}{4pt}},
}}
\newtcolorbox{promptenv}{promptstyle}
\newcommand{\promptbox}[1]{\begin{promptenv}#1\end{promptenv}}

\newenvironment{promptlist}
  {\begin{itemize}[leftmargin=1.2em, itemsep=3pt, topsep=2pt, parsep=0pt]}
  {\end{itemize}}

\newcommand{\jsonfence}{\textasciigrave\textasciigrave\textasciigrave}
\newcommand{\promptlabel}[1]{%
  \makebox[\linewidth][l]{\footnotesize\sffamily\bfseries #1}\par\vspace{1pt}}

\definecolor{cvprblue}{rgb}{0.21,0.49,0.74}
\usepackage[pagebackref,breaklinks,colorlinks,allcolors=cvprblue]{hyperref}

\usepackage{amssymb}
\usepackage{bbm}
\usepackage{booktabs}
\usepackage{xspace}
\usepackage{array}
\newcolumntype{H}{>{\setbox0=\hbox\bgroup}c<{\egroup}@{}}

\def\paperID{*****} % *** Enter the Paper ID here
\def\confName{CVPR}
\def\confYear{2026}

\newcommand{\mvi}{\textsc{MultiVENT 1.0}\xspace}
\newcommand{\mvii}{\textsc{MultiVENT 2.0}\xspace}

\newcommand{\mvraw}{\textsc{MultiVENT-Raw}\xspace}
\newcommand{\microv}{\textsc{MicroVENT}\xspace}

\newcommand{\rag}{\textsc{RAG}\xspace}
\newcommand{\oracle}{\textsc{Oracle}\xspace}
\newcommand{\infop}{Information P\xspace}
\newcommand{\infor}{Information R\xspace}
\newcommand{\infof}{Information $\text{F}_1$\xspace}
\newcommand{\citationp}{Citation P\xspace}
\newcommand{\citationr}{Citation R\xspace}
\newcommand{\citationf}{Citation $\text{F}_1$\xspace}

\newcommand{\mixed}{\textsc{Full}\xspace}
\newcommand{\raw}{\textsc{Core}\xspace}

\title{\mvraw: \\ A Benchmark for Retrieval and Reasoning over Raw Videos}

\author{Reno Kriz\textsuperscript{1,2} \quad David Etter\textsuperscript{1,2} \quad Alexander Martin\textsuperscript{2} \quad Cameron Carpenter\textsuperscript{2} \quad
Debashish Chakraborty\textsuperscript{1,2} \\ Hannah Recknor\textsuperscript{1,2} \quad Reihaneh Iranmanesh\textsuperscript{3} \quad Matthew Maciejewski\textsuperscript{1,2} \quad Kenton Murray\textsuperscript{1,2,5}\\ Eugene Yang\textsuperscript{1,2}
\quad Benjamin Van Durme\textsuperscript{1,2}
\quad Aaron Steven White\textsuperscript{4} \quad Andrew Yates\textsuperscript{1,2} \quad William Walden\textsuperscript{1,2} \\
\textsuperscript{1}\small{Human Language Technology Center of Excellence} \textsuperscript{2}\small{Johns Hopkins University}\\ 
\textsuperscript{3}\small{Georgetown University} 
\textsuperscript{4}\small{University of Rochester}
\textsuperscript{5}\small{George Mason University}\\
{\tt\small \{rekriz1,wwalden1\}@jh.edu}
}

\begin{document}
\maketitle
\begin{abstract}
Online information is increasingly consumed in video format. Much of this comes in the form of \emph{raw video}: continuous footage taken on a cell phone, with a hand-held camera, or via CCTV, which is then directly uploaded to social media platforms and content sharing services. Whereas professional or even amateur-edited footage tends to feature scripted speech, chyrons, graphics, and metadata that help contextualize its subject matter, raw video typically contains none of these things, making it a much more challenging medium for information retrieval and machine understanding. To facilitate progress in this domain, we release \mvraw, a multilingual collection of nearly 120,000 primarily raw videos (over 5,300 total hours), paired with 130 events and 222 event-centric queries, along with human-annotated video relevance judgments and human-extracted key facts for relevant videos. \mvraw supports both a \emph{retrieval} task---to identify videos in the collection relevant to a query event---and a \emph{generation} task---to summarize event-related videos into a coherent report for a target user. We benchmark strong baselines on \mvraw, showing both tasks to be challenging even for some of the latest multimodal models.\footnote{Code and data: \url{https://github.com/hltcoe/multivent-raw}.} 
\end{abstract}    

\section{Introduction}
\label{sec:intro}
Online information is increasingly conveyed in video. Beyond social media platforms and content sharing services---where video has long played a central role---even conventional print outlets now publish substantial amounts of video content online \cite{felps2026economist, robertson2026demand, steinberg2026ap}. Much of this content is, in whole or in part, \emph{raw} video: continuous footage filmed on cell phones, hand-held cameras, or CCTV, that is then uploaded, unedited, to an online platform. Whereas \emph{professional} or \emph{edited} footage contains large amounts of contextualizing information---such as narration, visual overlays, captions, and metadata---raw footage has far less of this content, and often none at all. Video of this kind thus poses unique challenges for multimodal retrieval and understanding: much of the context that is \emph{explicit} in edited video is instead \emph{implicit} in raw video, and must be inferred in order to match high-level user queries to relevant footage.

\begin{figure}
    \centering
    \includegraphics[width=\linewidth]{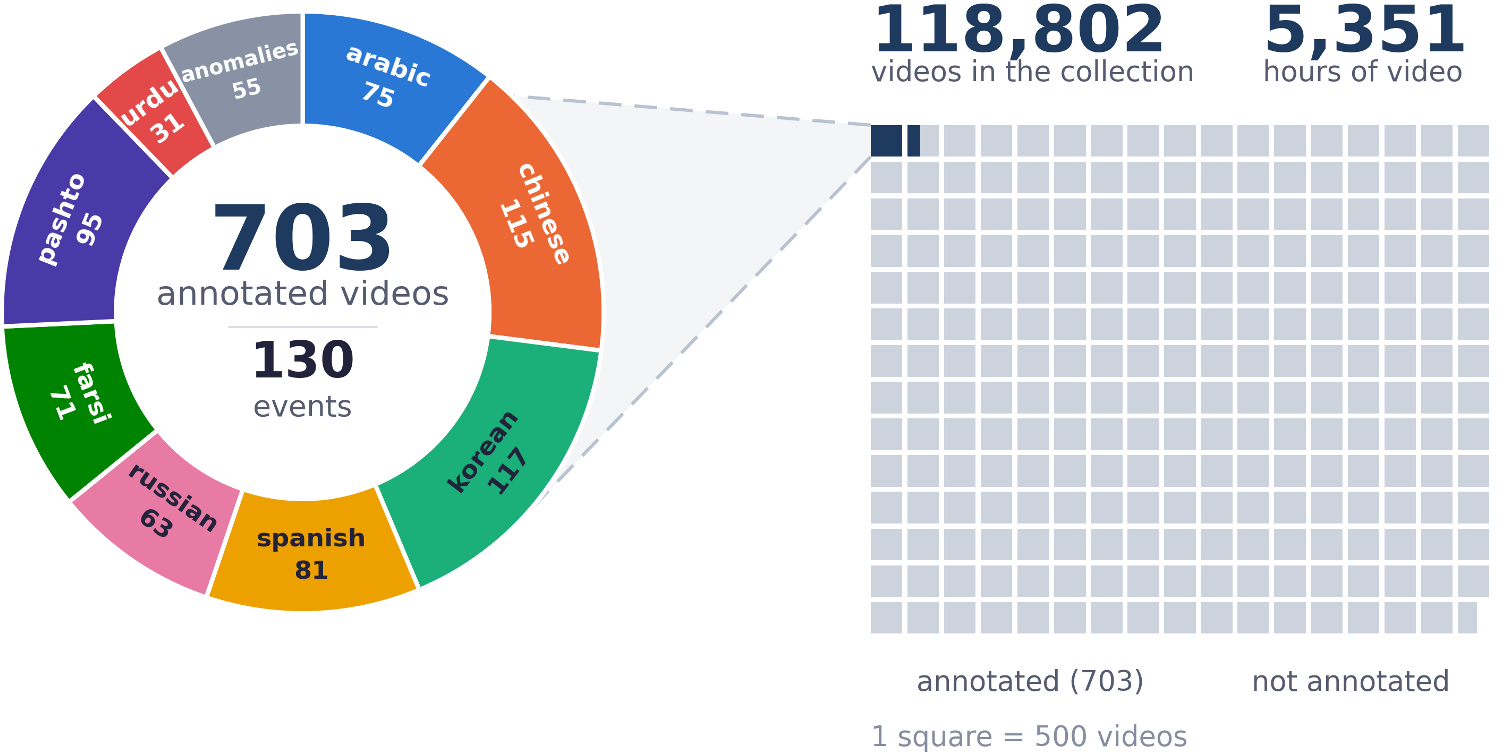}
    \caption{Summary statistics for \mvraw (\mixed).}
    \label{fig:fig1}
\end{figure}

Despite the centrality of raw video to digital platforms, benchmarks for video retrieval and understanding focus overwhelmingly on simple user queries over \emph{edited} videos \cite{sanders2023multivent,wang2024internvid,kriz2025multivent, chowdhury2026magnet, martin-etal-2026-wikivideo}. To remedy this, we introduce \mvraw---a large, multilingual collection of raw video for retrieval and generation against complex, event-centric queries. \mvraw spans 130 events, 222 queries, 8 languages, and nearly 120,000 videos, totaling over 5,300 hours of footage (\cref{fig:fig1}). Relevance judgments are annotated manually for all queries, and key facts, or \emph{claims}, are manually written for all relevant videos for each query. \mvraw topics cover a range of recent, real-world events, including natural disasters, protests, cultural celebrations, technological developments, and political incidents, among others. A further distinguishing feature of \mvraw is its coverage of \emph{anomalies}: a subset of topics focuses on unusual occurrences drawn from livestream cameras from around the world, which were sampled at regular intervals over the course of 11 months.
% events that are exceedingly unlikely to have received media coverage and thus to be encoded in models' parametric memory.
In summary, we:
\begin{enumerate}
    \item Release \mvraw, a large multilingual benchmark for retrieval and reasoning over raw videos, spanning 8 languages, 130 events, 222 queries, and nearly 120,000 videos. 
    \item Release \microv, a smaller dev set covering 23 events, useful for fast iteration on system development.
    \item Present baseline results on retrieval and generation, showing \mvraw to be a challenging benchmark for both tasks.
\end{enumerate}

\section{Related Work}
\label{sec:related-work}
% Our work builds upon multiple areas of video-based retrieval and reasoning.
\paragraph{Text-to-Video Retrieval: Benchmarks}
There are many benchmarks for video retrieval, spanning different video collections and catering to different use cases. Many notable benchmarks, such as \textsc{MSVD} \cite{chen2011collecting}, \textsc{Valor-32K} \cite{liu2024valor}, \textsc{InternVid} \cite{wang2024internvid}, and \textsc{MSR-VTT} \cite{xu2016msr}, focus exclusively on shorter, English-only video clips (averaging ${\leq}$15 seconds), often segmented from longer ones. Some benchmarks, such as \textsc{VATEX} \cite{wang2019vatex} and \textsc{Multi-HowTo100M} \cite{huang-etal-2021-multilingual}, do have \emph{queries} in multiple languages, but not \emph{videos} with multilingual speech content. All of the above benchmarks use video summaries or scene captions as queries. This has the advantage of scalability, but can result in bland, artificial queries that are not representative of real information needs. Lastly, the videos in these collections focus heavily (or even exclusively) on edited and professional content.

Addressing some of the points above, \citeauthor{sanders2023multivent} \cite{sanders2023multivent} introduce \mvi, a benchmark for event-centric video retrieval that covers 255 events (from 2015 to 2023) and ${\sim}$2,400 videos (both edited and raw) in multiple languages (Arabic, Chinese, English, Korean, and Russian). Queries are one- or two-sentence English event descriptions drawn from relevant text documents (e.g., news articles).

\mvii \cite{kriz2025multivent} expands \mvi to roughly 3,900 queries and 218,000 videos, covering the same event types and languages (+Spanish), and drawing on \textsc{InternVid} to source the additional videos. Queries in \mvii are more fine-grained than in \mvi, and ask about specific aspects of the target events (e.g., \emph{Changsha skyscraper fire casualties}) based on annotations from \cite{sanders-etal-2024-grounding}. This dataset also supported a shared task at the first Workshop on Multimodal Augmented Generation via Multimodal Retrieval (MAGMaR) \cite{magmar-ws-2025-1}.

\mvraw differs from all of the above in (1) limiting relevant videos to \emph{raw} content only; (2) expanding language coverage (adding Pashto, Farsi, and Urdu to the set of languages covered by \textsc{MultiVENT 1.0} and \textsc{2.0}); and (3) using realistic, multi-sentence analytic queries rather than short questions or event descriptions.

\paragraph{Text-to-Video Retrieval: Methods}
Text-to-video retrieval systems typically follow a two-stage \emph{retrieve-then-rerank} pipeline that pairs an efficient first-stage retriever with a more expressive reranker. Most first-stage retrievers are dense bi-encoders that embed queries and videos as single vectors
\cite{ma2025tevatron, xu2025omniembednemotronunifiedmultimodalretrieval, chen-etal-2026-e5, zhang2025gmeimprovinguniversalmultimodal, hönicke2026jinaembeddingsv5omnigeometrypreservingembeddingslocked, li2026qwen3vlembeddingqwen3vlrerankerunifiedframework} or multiple vectors \cite{reddy2025videocolbert, qin2026multivectorindexcompressionmodality}. In the second stage, \emph{video-native reranking} models rescore an initial candidate list by reasoning over video content directly \cite{skow2026rankvideoreasoningrerankingtexttovideo, li2026qwen3vlembeddingqwen3vlrerankerunifiedframework}. As an alternative to encoding video directly, some systems derive video representations from extracted or generated text \cite{wu2023cap4videoauxiliarycaptionstextvideo, degenaro-etal-2025-fortify}, or encode each modality independently and fuse the resulting rankings \cite{samuel2025mmmorrf}.

\paragraph{Multi-Video Retrieval-Augmented Generation} The report generation task that \mvraw supports is a \emph{multi}-video retrieval-augmented generation (RAG) task. While there is substantial work on \emph{single-}video RAG \cite{yuan2025memoryenhanced, luo2026video, fu-etal-2026-videostir, lee2026rethinking}, work in the multi-video setting has only recently begun to receive meaningful attention---likely due not only to high compute costs, but also to the lack of native support for multi-video processing even in the most recent multimodal models. \citeauthor{ren2026videorag} \cite{ren2026videorag} introduce the VideoRAG framework, which enables information synthesis over multiple videos via a knowledge graph constructed from entities and relations attested in individual videos. \citeauthor{tevissen2024towards} \cite{tevissen2024towards} perform multi-video RAG entirely in text, indexing videos on speech transcripts and visual metadata, and retrieving over this index to answer simple questions about an archive of NASA footage. Lastly, the \emph{second} MAGMaR Workshop \cite{magmar-2026-main} organized a shared task around the \textsc{WikiVideo} benchmark from \cite{martin-etal-2026-wikivideo}, which requires generating a Wikipedia-style article about a query event based on videos retrieved from a subset of the \textsc{MultiVENT 2.0} dataset. In \S\ref{sec:experiments}, we present results both with the baseline system for this shared task (CAG \cite{martin-etal-2026-wikivideo}) and with two top participant systems (TRACE \cite{yan2026trace} and MARQUIS \cite{chakraborty-etal-2026-marquis}).

\section{Data Collection}
\label{sec:data-collection}
\begin{figure*}
    \centering
    \includegraphics[width=0.9\linewidth]{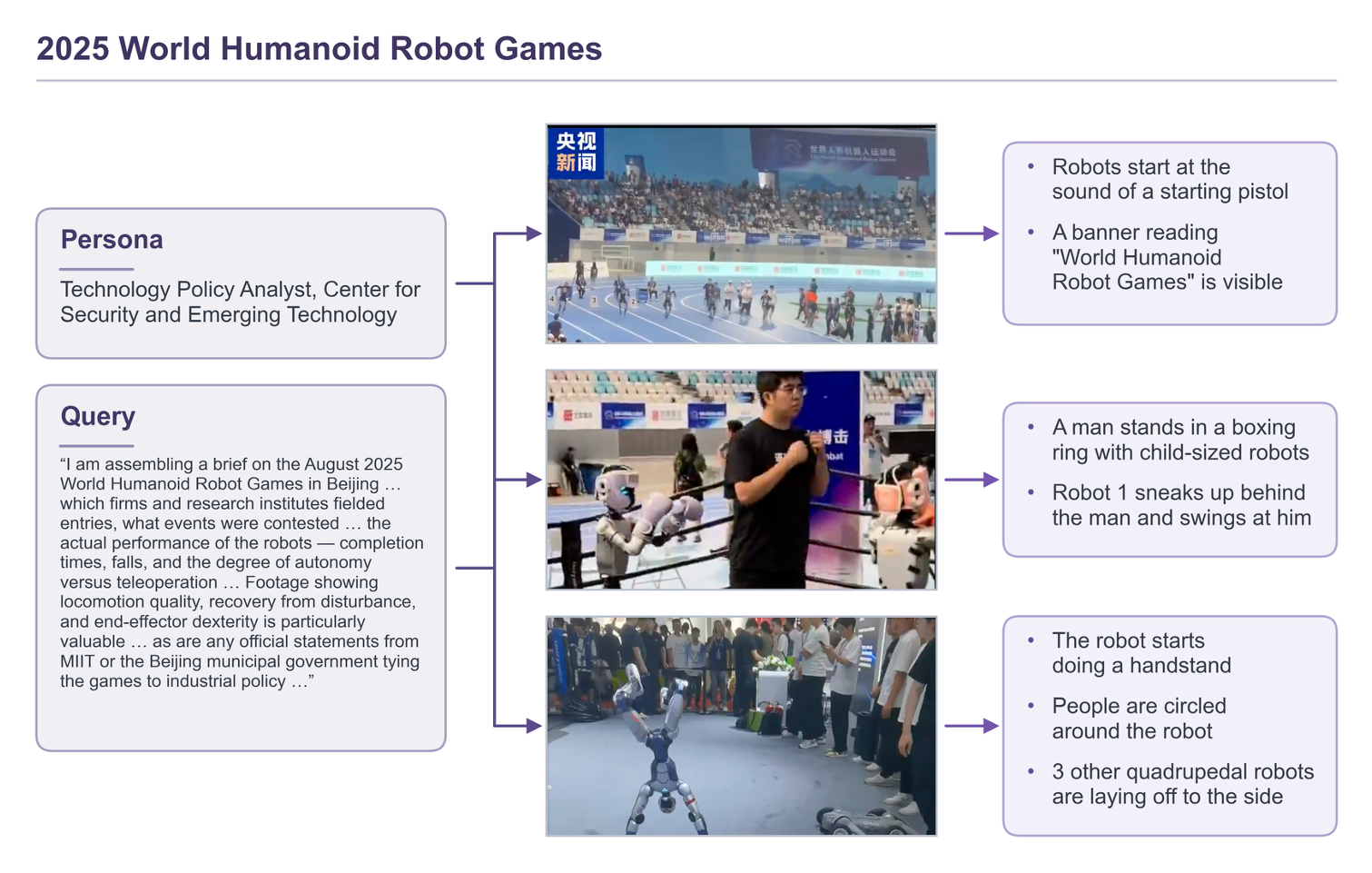}
    \caption{An example of a persona title (description omitted), a query, and (a sample of) claims from a \mvraw video about the 2025 World Humanoid Robot Games in Beijing.\vspace{-5mm}}
    \label{fig:example-event}
\end{figure*}

\mvraw contains \emph{raw} video. We define this as continuous footage that is effectively unscripted and unedited---excluding video with scripted speech, narration, music tracks, graphical or text overlays (e.g., chyrons, captions), or multiple scenes. Thus, news broadcasts, amateur-edited videos, and compilations are all filtered out.\footnote{Minimal graphical and text overlays are permitted in select cases (e.g., channel logos in the corner of the screen or timestamps from livestream cameras) when the footage is otherwise fully raw.}

\subsection{Topic and Video Curation}
\label{subsec:data-collection-topic-curation}
Data collection for \mvraw began with topic curation. The curation process differed for our \textbf{primary events}, drawn mainly from cell phone and hand-held camera footage, and our \textbf{anomaly events}, drawn entirely from livestreams and CCTV cameras.

\paragraph{Primary Events} We sought to make \mvraw a multilingual benchmark, focusing on eight target languages---Arabic, Chinese, Farsi, Korean, Pashto, Russian, Spanish, and Urdu---chosen based on the linguistic competencies of available annotators (see \S\ref{subsec:data-collection-claim-annotation}). For each target language, we searched the web for recent events in countries where that language is widely spoken, though we did not \emph{require} videos depicting these events to contain speech. Further, although we placed no hard restrictions on the types of events to curate, we prioritized those that seemed likely to have been documented in raw video footage.\footnote{Thus, this process selects against types of events likely to receive coverage only in highly edited form (e.g., political debates, diplomatic talks, or financial incidents).} Local and recent events were also prioritized to minimize the likelihood and extent of their coverage in the training data of current multimodal models. Our search involved a mix of manual discovery on news sites and social media platforms (YouTube, TikTok, and X) and automatic discovery via a Claude skill that we designed for this purpose \cite{anthropic2026agentskills}. In both cases, an event was retained only if we were able to find \emph{multiple} raw videos depicting it.

\paragraph{Anomaly Events} Anomaly events are occurrences captured on a livestream that are somehow unusual compared to the typical activity on that stream. Relative to our primary events, they add two further challenges. First, since they are sourced from individual cameras, they are even less likely to have received media coverage and thus to have representation in models' training data. Second, the camera activity corresponding to the event of interest is typically embedded within tens or hundreds of minutes of similar footage (vs.\ only several minutes for primary events).

For anomaly events, we curated topics from a pool of 322 public livestream cameras, which we sampled at regular intervals over 11 months (June 2025 through April 2026). The resulting 24,459 clips come from cameras in over 20 countries and depict diverse locations including traffic intersections, parks, rail lines, harbors, airports, and volcanoes. Our annotators identified anomalies by first reviewing each camera's footage to establish a normal activity profile, and then selecting clips that deviate from that profile. We dropped cameras where more than 4\% of clips were judged anomalous, and each remaining camera with at least one anomaly became a topic. Topics cover a diverse range of events---e.g., a volcanic eruption, a team of fire trucks driving onto an airport runway, and a car parking on the grass in a park in the middle of the night.

\paragraph{Video Relevance Annotation}
All primary event videos and all anomalous clips identified above are deemed \emph{relevant} to the corresponding event. For anomaly events, all remaining (irrelevant) clips are retained in the collection as hard negatives. For primary events, we mined hard negatives by retrieving videos from a larger candidate collection, sourced primarily from YouTube, with additional videos from X and TikTok. We retrieved candidates using Qwen3-VL-Embedding (8B) \cite{li2026qwen3vlembeddingqwen3vlrerankerunifiedframework} representations of visual keyframes and Qwen3-Embedding (8B) \cite{zhang2025qwen3embeddingadvancingtext} representations of speech transcripts and OCR text, produced by Qwen3-ASR (1.7B) \cite{shi2026qwen3asrtechnicalreport} and PaddleOCR-VL-1.6 \cite{zhang2026paddleocrvl16expandingfrontierdocument}, respectively. For each query, we ranked candidates by embedding similarity and retained high-scoring clips after excluding videos that were already judged relevant to that query. 

% This procedure selects potentially confusable distractors, although clips without positive judgments may still be relevant and can receive updated labels in the subsequent annotation round. 
%\todo{Deb: discuss the ``larger candidate collection'' and the models used for retrieval. Also clarify the last sentence in this paragraph.}

% BELOW: which baselines?
A second round of manual relevance judgments was conducted following training of retrieval baselines on this collection, with annotations collected on the top 20 relevant videos for each query as judged by these baselines (see $\S\ref{sec:experiments}$). These relevance judgments were then added to the originals to yield the final set of judgments for the collection.

\paragraph{Summary Statistics} In total, we identified 703 relevant videos covering 130 events (96 primary, 34 anomaly), of which 102 events occurred in 2025--2026. \cref{fig:fig1} shows the breakdown of relevant videos by language. The remaining 118,099 videos in the collection are hard negatives.

\begin{figure}
    \centering
    \includegraphics[width=\linewidth]{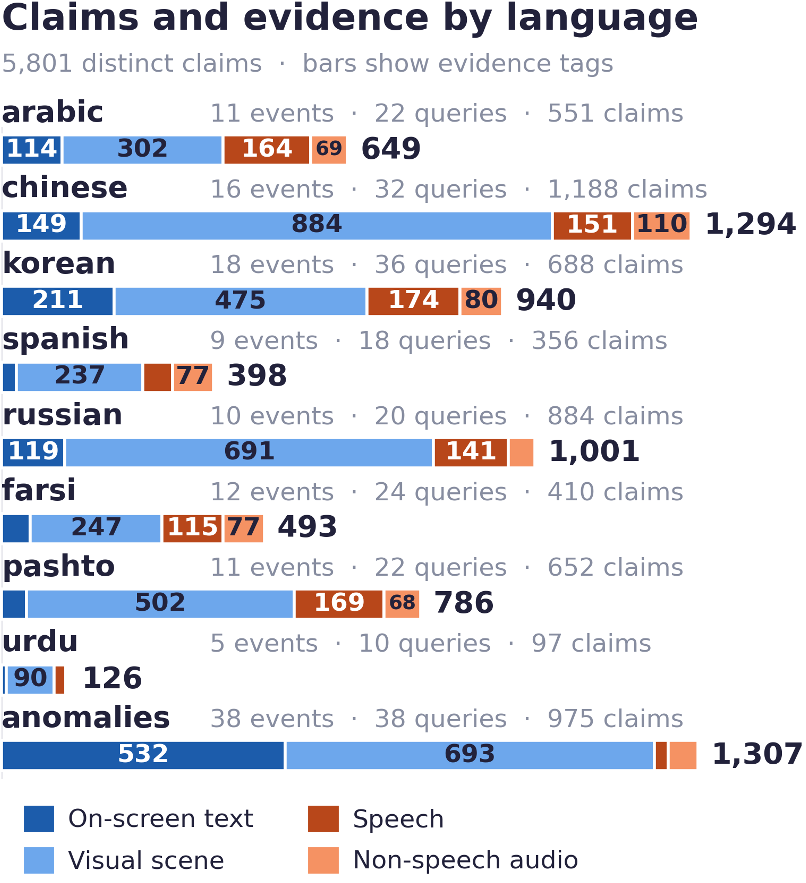}
    \caption{Breakdown of \mvraw events, queries, and (original) claims by language. Claims are further broken down by type of evidential support (see \S\ref{subsec:data-collection-claim-annotation}). Claims may be counted multiple times if supported by multiple modalities, hence the difference between the bolded numbers and the total claim counts.}
    \label{fig:claim-stats}
\end{figure}

\subsection{Query and Persona Development}
\label{subsec:data-collection-query-and-persona-development} For each topic, we sought to develop queries that capture complex, realistic information needs. More specifically, we wanted queries that were more analytical than those in related work (e.g., \cite{sanders2023multivent, kriz2025multivent}), that would require reasoning over \emph{multiple} videos, and that might plausibly have been asked by someone affected by the target event.

\paragraph{Personas} To aid in query development, we first crafted a profile, or \emph{persona}, describing a hypothetical person who would have a stake in the event and its consequences. Each persona consists of a job title and description (see \cref{fig:example-event}), which provide broader context for the query, and may be used alongside it for both retrieval and generation.

\paragraph{Queries} There are at least two types of information need a user might have about a target event. On the one hand, they may be interested in finding \emph{any and all} information about the event that is relevant to their responsibilities, requiring a system to find and summarize a \emph{broad} set of related videos. On the other hand, the user may have a more specific set of questions about what happened, requiring a system to find \emph{particular} videos that address these questions or else determine that no answer can be found. Accordingly, for each primary event, we construct two queries---one for each type of need. For both, a small initial query set was manually written by the authors. These queries then informed the creation of a Claude skill that was used to write the remainder.

\emph{Broad} queries capture the first use case and are written based on the persona and the claim annotations (see \S\ref{subsec:data-collection-claim-annotation}) from all relevant videos for the target event. We use the claim annotations so that the query conforms to the set of event-related information \emph{actually attested} in the collection, consistent with the summarization-style objective.

\emph{Narrow} queries cover the second use case and are written based only on the persona and the titles of relevant videos. In this way, queries are informed at a high level by relevant content attested in the collection, but are also free to pose questions for which the collection may supply no answers.

Since anomaly detection fits a summarization-style objective---where \emph{any and all} anomalous incidents are to be flagged---we write only broad queries for the anomaly events. In total, we obtain 222 queries.

\subsection{Claim Annotation}
\label{subsec:data-collection-claim-annotation}
To enable reliable evaluation of our generation task (\S\ref{subsec:tasks-report-generation}), we annotated key facts, or \emph{claims}, for each relevant video. Annotators were given a target persona, asked to watch a video, and then asked to write claims (in English) that capture important information from the video that the target persona would likely want to know. Annotators further provided confidence scores (0--100) for each claim\footnote{In practice, annotators were asked not to write claims for which their confidence score was ${<}$50, as this would suggest they believe the claim to be more likely false than true.} and indicated which of the video's modalities provided evidence for the claim, with the option to select multiple: (a) the visual frames, excluding text; (b) the on-screen text; (c) the speech; or (d) the non-speech audio. 

Linguists with expertise in each of our target languages were recruited to perform the annotation, using the Flexible Ontology Visual Event Analyzer (FOVEA) annotation tool to do so \cite{white2026fovea}. All annotators were given a tutorial as well as a detailed set of written instructions, and completed a practice annotation that was reviewed and discussed with the authors. Annotators were also permitted to ask clarifying questions throughout the task. In total, annotators wrote roughly 5,800 claims across the 703 relevant videos in \mvraw, averaging 8.25 claims per video.

\subsection{Quality Control and Postprocessing}
\label{subsec:postprocessing}

\paragraph{Videos}
Although annotated videos were manually verified as raw, \emph{irrelevant} videos in the collection (obtained via hard negative mining) could not be verified in the same way, given their large number.

Accordingly, we developed a classifier to distinguish raw from edited videos. To do so, we relied on the raw/edited labels provided for the \textsc{MultiVENT 2.0} test set \cite{kriz2025multivent}, dividing this randomly into train (85\%) and dev (15\%) splits. We assessed an array of different classifier designs, the best of which was an ensemble system that aggregates predictions from two zero-shot prompted VLMs, two fine-tuned generative VLM classifiers, and a classifier based on FFmpeg features \cite{ffmpeg}.\footnote{Appendix \ref{append:data-details} has further details on this classifier.} This classifier achieved an $\text{F}_1$ of 91.4\% on the dev set, indicating high reliability. The classifier was then run over all irrelevant videos in the collection.

% 75,147 total videos in filtered set - 703 relevant = 74,444 filtered and irrelevant
% Something to be resolved

Since raw video retrieval use cases include settings where edited videos are present in the collection and settings where they are not, we define two splits of \mvraw: \mixed, which contains all 118,802 original videos and \raw, which removes all irrelevant videos labeled as edited by our classifier, leaving 59,196.
% We present baseline results on both splits in \S\ref{sec:experiments}.

\paragraph{Claims} Claim postprocessing involved three tasks, all implemented as Claude skills that drew on insights from an initial manual review of annotated claims.

The first task was to ensure that all claims were unique, atomic (short, non-compound), and self-contained (e.g., with ambiguous references resolved). Claims that could not be rewritten to satisfy these criteria were discarded.

The second task was to filter the original set of claims (\emph{original claims}) into subsets depending on their relevance to the target event (\emph{event claims}), persona (\emph{persona claims}), and query (\emph{query claims}). Although annotators were instructed to write persona claims (\S\ref{subsec:data-collection-claim-annotation}), in practice we found that some annotated claims were overly general (i.e., were more properly classed as event claims) and that only a subset were further relevant to the query (i.e., were query claims). Here, we assume that $\{\textit{query claims}\} \subseteq \{\textit{persona claims}\} \subseteq \{\textit{event claims}\} \subseteq \{\textit{original claims}\}$.

The third task was to augment query claims with \emph{negative claims}---claims asserting that \emph{no} answer to some facet of the query is provided by the collection. We write one such claim for each facet for which this is the case. Note that, because negative claims are not grounded in any video, they lack modality support annotations (in contrast to the other \emph{positive} claims).

\subsection{\textbf{\microv}}
\label{subsec:data-collection-microvent}
To facilitate more rapid system development, and to allow all of \mvraw to serve as a test set, we release alongside it a dev set that we call \microv. \microv consists of 23 events, 31 queries (23 biased, 8 unbiased), and 933 total videos (272 annotated). Like \mvraw, \microv covers both primary events (18) and anomaly events (6), curated and annotated in the same fashion described above. Unlike \mvraw, however, 10 of the primary events in \microv are drawn from events in \textsc{MultiVENT 2.0}, which covers similar types of events to \mvraw (see \S\ref{sec:related-work}). As with the \mixed split of \mvraw, we do not filter out edited videos from the \microv collection.

% \todo{Describe \microv, including summary statistics. Discuss similarities and differences in data collection relative to \mvraw. We should NOT report any results on \mvraw in the main text.}

\section{Tasks}
\label{sec:tasks}
\begin{table*}[t]
\centering
\small
\begin{tabular}{lrrrrrrrr}
\toprule
& \multicolumn{4}{c}{\bf\mixed} & \multicolumn{4}{c}{\bf\raw} \\
\cmidrule(lr){2-5}\cmidrule(lr){6-9}
\bf System & \bf nDCG@10 & \bf nDCG@20 & \bf R@20 & \bf R@100 & \bf nDCG@10 & \bf nDCG@20 & \bf R@20 & \bf R@100 \\
\midrule
{ColQwen Omni (3B)} & 23.3 & 23.9 & 25.1 & 39.2 & 31.9 & 32.9 & 33.1 & 45.8 \\
{Qwen3-VL (8B)} & 26.7 & 27.8 & 29.2 & 45.6 & 32.5 & 33.3 & 34.3 & 48.8 \\
{OmniEmbed (7B)} & 18.0 & 18.7 & 20.5 & 34.9 & 21.2 & 22.1 & 22.2 & 34.1 \\
{OmniEmbed-mv (7B)} & 47.2 & 48.9 & 51.6 & \textbf{71.8} & \textbf{49.7} & 50.8 & 51.9 & \textbf{67.2} \\
\ \ + {RankVideo (8B)} & 48.5 & 51.7 & 57.2 & \textbf{71.8} & 49.3 & \textbf{51.1} & \textbf{54.5} & \textbf{67.2} \\
\ \ + {Gemma 4 (12B)} & \textbf{51.5} & \textbf{54.4} & \textbf{57.3} & \textbf{71.8} & 47.9 & \textbf{51.1} & 53.3 & \textbf{67.2} \\
\bottomrule
\end{tabular}
\caption{Retrieval baselines on \mvraw.
The best value in each column is bolded. Results with RankVideo and Gemma 4 are reranking results on  OmniEmbed-mv outputs. All other results are first-stage retrieval results.}
\label{tab:mvraw-retrieval-baselines}
\end{table*}

\subsection{Video Retrieval}
\label{subsec:tasks-video-retrieval}
% The \mvraw retrieval task is identical to the one proposed in \mvi.
Given a query $q_e$ about an event $e$, an ideal retrieval system should return a ranked list of $n$ videos from the collection, $L_{q_e} = [v_1, \ldots, v_n]$, where all videos relevant to $q_e$ are ranked higher than all irrelevant videos. In \S\ref{sec:experiments}, we use the annotated queries as $q_e$, excluding the personas.
% \todo{Will: verify this is true}.
%Since we use binary relevance judgments, relative ranking among relevant videos is unimportant.
We report recall@$k$ (R@$k$) for $k\in\{20,100\}$ (where $k \leq n$), defined as the fraction of all $q_e$-relevant videos in the collection that appear among the top $k$ of $L_{q_e}$, averaged over all $q_e$. We also report nDCG@$k$ for $k\in\{10,20\}$, which measures the degree to which $L_{q_e}$ approximates a perfect rank order.

\subsection{Report Generation}
\label{subsec:tasks-report-generation}
Our report generation task is similar to the event-centric article generation task from the \textsc{WikiVideo} benchmark \cite{martin-etal-2026-wikivideo}. Given $e$, $q_e$, and $L_{q_e}$, a system must produce a report that addresses $q_e$ as completely and accurately as possible, with citations to one or more supporting videos $v \in L_{q_e}$ appended following each report sentence that requires evidential support.\footnote{In our experiments, we assume that \emph{every} report sentence requires evidential support.} We consider two settings: a \rag setting, where $L_{q_e}$ is provided by a retrieval system (\S\ref{subsec:tasks-video-retrieval}) and an \oracle setting, where $L_{q_e}$ consists of the ideal ranked list for $q_e$ ($L^*_{q_e}$). Whereas the former setting represents our primary task of interest, the latter allows us to assess upper bounds on generation performance, independent of retrieval quality. We evaluate predicted report quality with MiRAGE \cite{martin2026seeingmirageevaluatingmultimodal}, a suite of metrics for multimodal RAG tasks that measure the degree to which a report both (a) recovers facts known to be attested by relevant videos (information precision, recall, and F1 score) and (b) provides appropriate citation for its claims (citation precision, recall, and F1 score).\footnote{Appendix \ref{append:eval-details} has details on these metrics.}

\section{Experiments}
\label{sec:experiments}
% requires \usepackage{multirow}
\begin{table*}[t]
\centering
\small
\setlength{\tabcolsep}{4pt}
\begin{tabular}{llrrrrrrrrrrrr}
\toprule
& & \multicolumn{6}{c}{\bf\mixed} & \multicolumn{6}{c}{\bf\raw} \\
\cmidrule(lr){3-8}\cmidrule(lr){9-14}
& & \multicolumn{3}{c}{\bf Information} & \multicolumn{3}{c}{\bf Citation} & \multicolumn{3}{c}{\bf Information} & \multicolumn{3}{c}{\bf Citation} \\
\cmidrule(lr){3-5}\cmidrule(lr){6-8}\cmidrule(lr){9-11}\cmidrule(lr){12-14}
\bf Retriever & \bf System & \bf P & \bf R & \bf F1 & \bf P & \bf R & \bf F1 & \bf P & \bf R & \bf F1 & \bf P & \bf R & \bf F1 \\
\midrule
\multirow{6}{*}{OmniEmbed (7B)}
 & CAG                             & 9.5 & 22.3 & 10.7 & 2.1 & 3.1 & \textbf{2.1} & 8.7 & 22.1 & 10.3 & 2.5 & 4.5 & 2.4 \\
 & MARQUIS\textsubscript{CAG} & 15.3 & 22.3 & 14.6 & 1.9 & 3.0 & 1.6 & 15.9 & 21.5 & 14.4 & 2.9 & 4.6 & 2.4 \\
 & MARQUIS\textsubscript{bullet}   & 20.2 & 23.4 & 16.1 & 2.5 & 4.0 & 2.0 & 20.1 & 22.0 & 16.3 & 4.0 & 6.2 & \textbf{3.2} \\
 & MARQUIS\textsubscript{ginger}   & 15.7 & 21.2 & 13.5 & 2.1 & 2.4 & 1.4 & 17.9 & 20.6 & 14.6 & 2.6 & 3.9 & 2.1 \\
 & TRACE\textsubscript{lite}       & 17.2 & 31.7 & \textbf{19.7} & 1.2 & 5.1 & 1.7 & 16.2 & 28.4 & \textbf{18.0} & 2.1 & 7.8 & 2.9 \\
 & TRACE\textsubscript{full}       & 15.6 & 28.0 & 17.4 & 1.4 & 4.5 & 1.9 & 15.2 & 24.6 & 16.4 & 2.4 & 7.2 & 2.9 \\
\cmidrule(lr){1-14}
\multirow{6}{*}{OmniEmbed-mv (7B)}
 & CAG                             & 9.6 & 30.6 & 12.6 & 3.4 & 10.9 & 4.1 & 10.6 & 31.5 & 13.1 & 4.6 & 13.0 & 5.1 \\
 & MARQUIS\textsubscript{CAG} & 17.6 & 26.8 & 17.2 & 5.0 & 9.1 & 4.6 & 19.6 & 27.6 & 18.6 & 6.4 & 10.7 & 5.2 \\
 & MARQUIS\textsubscript{bullet}   & 19.7 & 28.0 & 18.4 & 5.7 & 12.4 & \textbf{5.4} & 22.6 & 27.2 & 19.2 & 8.3 & 13.7 & \textbf{7.1} \\
 & MARQUIS\textsubscript{ginger}   & 18.0 & 25.2 & 16.0 & 4.5 & 7.6 & 3.4 & 20.4 & 25.6 & 17.1 & 5.5 & 9.1 & 4.0 \\
 & TRACE\textsubscript{lite}       & 20.8 & 40.1 & \textbf{24.6} & 3.6 & 15.9 & 5.1 & 21.2 & 40.9 & \textbf{25.2} & 5.1 & 19.8 & 7.1 \\
 & TRACE\textsubscript{full}       & 19.2 & 35.7 & 22.3 & 3.6 & 14.3 & 4.9 & 20.0 & 34.7 & 22.6 & 5.2 & 17.5 & 6.7 \\
\cmidrule(lr){1-14}
\multirow{6}{*}{\ \ + RankVideo (8B)}
 & CAG                             & 10.9 & 31.9 & 13.6 & 3.9 & 11.9 & 4.8 & 11.0 & 32.0 & 13.4 & 4.8 & 13.8 & 5.6 \\
 & MARQUIS\textsubscript{CAG} & 20.5 & 28.2 & 19.9 & 5.5 & 10.6 & 4.9 & 21.1 & 27.1 & 19.9 & 5.9 & 10.5 & 5.0 \\
 & MARQUIS\textsubscript{bullet}   & 22.5 & 29.5 & 20.3 & 7.0 & 14.6 & \textbf{6.9} & 24.6 & 29.3 & 21.4 & 9.5 & 15.9 & \textbf{8.3} \\
 & MARQUIS\textsubscript{ginger}   & 17.1 & 25.6 & 15.8 & 4.6 & 8.9 & 3.4 & 19.4 & 25.7 & 16.3 & 7.0 & 10.6 & 4.9 \\
 & TRACE\textsubscript{lite}       & 22.1 & 41.6 & \textbf{26.2} & 4.0 & 18.6 & 5.7 & 22.0 & 40.6 & \textbf{25.9} & 5.0 & 21.1 & 7.1 \\
 & TRACE\textsubscript{full}       & 20.8 & 37.9 & 24.3 & 4.2 & 16.6 & 5.7 & 21.6 & 36.2 & 24.4 & 5.1 & 18.5 & 6.7 \\
\bottomrule
\end{tabular}
\caption{MiRAGE metric results (``reference'' variant, using Qwen3.5-27B as judge) on the \mixed and \raw splits of \mvraw, given the top $k=10$ videos from three different retrieval systems. Retriever names follow Table~\ref{tab:mvraw-retrieval-baselines}, which reports the corresponding retrieval effectiveness.
%All conditions share one query set, one reference and one authoritative collection, so cells are directly comparable.
The best F1 in each column within a retriever block is bolded.}
\label{tab:rag-qwen-reference-k10}
\end{table*}

We present baseline results on both the \mixed and \raw splits of \mvraw, for both the retrieval (\S\ref{subsec:tasks-video-retrieval}) and report generation tasks (\S\ref{subsec:tasks-report-generation}). For both, we chunk all videos into segments of at most 5 minutes, yielding 143,288 segments for \mixed and 75,147 for \raw. Thus, we index at the segment level for retrieval and we provide segments (rather than full videos) as VLM inputs.

\subsection{Retrieval}
\label{subsec:baselines-retrieval}

\paragraph{Baselines} For first-stage retrieval, we consider the following baseline methods, constructing separate indexes for the \mixed and \raw splits in each case:
\begin{enumerate}
    \item \textbf{ColQwen Omni (3B) \cite{faysse2025colpali, faysse2025colqwenomni}}. Multi-vector dense retrieval following the ColBERT late-interaction architecture \cite{khattab2020colbert}; uses Qwen2.5-Omni (3B) \cite{xu2025qwen25omni} fine-tuned on the ColPali training set \cite{faysse2025colpali}.
    \item \textbf{Qwen3-VL-Embedding (8B) \cite{li2026qwen3vlembeddingqwen3vlrerankerunifiedframework}}. Single-vector dense retrieval on Qwen3-VL embeddings.
    \item \textbf{OmniEmbed (7B) \cite{ma2025tevatron}}. Single-vector dense retrieval on OmniEmbed (Tevatron-Omni) embeddings. OmniEmbed is fine-tuned from Qwen2.5-Omni (7B) \cite{xu2025qwen25omni} on a suite of multimodal datasets.
    \item \textbf{OmniEmbed-mv (7B) \cite{ma2025tevatron}}. Single-vector dense retrieval on embeddings from a version of OmniEmbed fine-tuned on the \textsc{MultiVENT 2.0} train split.
    %\footnote{ColQwen Omni was introduced as a capable video retriever despite not having been trained on any audio or video data.}
    % \item \todo{Video LSR?} -> not a baseline
\end{enumerate}
We additionally use two baseline rerankers to rerank the top 100 videos from OmniEmbed-mv, which we find to be our strongest first-stage retrieval baseline.
%from \todo{which first-stage outputs?}:
\begin{enumerate}
    \item \textbf{RankVideo \citep{skow2026rankvideoreasoningrerankingtexttovideo}}. A video-native pointwise reranker initialized from Qwen3-VL (8B) and trained on videos from \textsc{MultiVENT 2.0}.
    \item \textbf{Gemma 4 (12B) \citep{team2026gemma}}. Zero-shot
pointwise reranking with off-the-shelf Gemma 4 (12B).\footnote{See Appendix~\ref{append:retrieval} for the prompt and additional details.}
\end{enumerate}

% SOMETHING LIKE: WE EVALUTE ALPHA NDCG BECAUSE \cite{samuel2026relevance}

% \begin{table*}[t]
% \centering
% \small
% \begin{tabular}{lrrrrrrrr}
% \toprule
% & \multicolumn{4}{c}{\bf\mixed} & \multicolumn{4}{c}{\bf\raw} \\
% \cmidrule(lr){2-5}\cmidrule(lr){6-9}
% \bf System & \bf nDCG@10 & \bf nDCG@20 & \bf R@20 & \bf R@100 & \bf nDCG@10 & \bf nDCG@20 & \bf R@20 & \bf R@100 \\
% \midrule
% {ColQwen Omni (3B)} & 23.3 & 23.9 & 25.1 & 39.2 & 31.9 & 32.9 & 33.1 & 45.8 \\
% {Qwen3-VL (8B)} & 26.7 & 27.8 & 29.2 & 45.6 & 32.5 & 33.3 & 34.3 & 48.8 \\
% {OmniEmbed (7B)} & 18.0 & 18.7 & 20.5 & 34.9 & 21.2 & 22.1 & 22.2 & 34.1 \\
% {OmniEmbed-mv (7B)} & 47.2 & 48.9 & 51.6 & \textbf{71.8} & \textbf{49.7} & 50.8 & 51.9 & \textbf{67.2} \\
% \ \ + {RankVideo (8B)} & 48.5 & 51.7 & 57.2 & \textbf{71.8} & 49.3 & \textbf{51.1} & \textbf{54.5} & \textbf{67.2} \\
% \ \ + {Gemma 4 (12B)} & \textbf{51.5} & \textbf{54.4} & \textbf{57.3} & \textbf{71.8} & 47.9 & \textbf{51.1} & 53.3 & \textbf{67.2} \\
% \bottomrule
% \end{tabular}
% \caption{Retrieval baselines on \mvraw.
% The best value in each column is bolded. Results with RankVideo and Gemma 4 are reranking results on  OmniEmbed-mv outputs. All other results are first-stage retrieval results.}
% \label{tab:mvraw-retrieval-baselines}
% \end{table*}

\paragraph{Results}
Retrieval results are shown in \autoref{tab:mvraw-retrieval-baselines}.
Among the first-stage retrieval methods, OmniEmbed-mv consistently achieves best results by a wide margin across all metrics, testifying to the value of domain-specific fine-tuning.
Qwen3-VL-Embedding reliably outperforms ColQwen Omni by a small margin, and both perform substantially better than the OmniEmbed model without \textsc{MultiVENT 2.0} fine-tuning. Comparing the same models across splits (\mixed vs.\ \raw), we find that \mixed is more difficult across the board---likely due to the semantically rich \emph{edited} hard negatives that are present in this split but missing from \raw.

Reranking consistently improves performance on the \mixed split, with Gemma 4 achieving the best results.
On the \raw split, reranking improves R@20 but does not have a consistent effect on nDCG.
Overall, these results illustrate that general-purpose video retrieval models struggle to effectively retrieve raw videos---particularly when similar but irrelevant \emph{edited} videos are also present---indicating the importance of evaluating models in this domain.

\subsection{Report Generation}
\label{subsec:baselines-generation}

\paragraph{Baselines} We consider three baseline systems for report generation, drawn from the ``grounded generation'' shared task at the second Workshop on Multimodal Augmented Generation via Multimodal Retrieval (MAGMaR) \citep{magmar-2026-main,martin-etal-2026-findings}:

\begin{enumerate}
    \item \textbf{Collaborative Article Generation (CAG) \citep{martin-etal-2026-wikivideo}}. The workshop's baseline system, CAG uses a VLM to produce a summary for each of the top-$k$ videos $v_1, \ldots, v_k \in L_{q_e}$. The VLM then condenses all $k$ video-level summaries into a final report with embedded citations. We use the CAG-0 variant, which generates a single summary per video, rather than the CAG-$N$ variants, which iteratively refine these summaries.
    \item \textbf{TRACE \citep{yan2026trace}}. One of the top-performing systems on the shared task, TRACE constructs rich, multimodal timelines for each video---consisting of timestamped features and a query-conditioned summary---from which claims are extracted and clustered to produce statements in the final report with accompanying citations. We use their full (features: ASR, OCR, object detection; key frame: uniform, filtered, and guided) and lite (features: ASR; key frame: uniform) variants, denoted as TRACE\textsubscript{[variant]}.
    \item \textbf{MARQUIS \citep{chakraborty-etal-2026-marquis}}. Another top-performing system, MARQUIS performs different types of extraction over each retrieved video (claim extraction, answers to event-related questions) and applies different aggregation techniques over retrieved evidence to produce the cited article. We use their Bullet (reports as bulleted lists of facts), Ginger \cite[following][]{lajewska2025gingergroundedinformationnuggetbased}, and CAG \cite[following][]{martin-etal-2026-wikivideo} variants, which we denote as MARQUIS$_{\text{[variant]}}$. 
\end{enumerate}

To maximize comparability across the three baselines, we use the same backbones for text (Qwen3-30B-A3B-Instruct-2507 \cite{yang2025qwen3}) and for vision (Qwen3-VL-30B-A3B-Instruct \cite{bai2025qwen3}) across systems. Here, we present results in the \textsc{RAG} setting, using top-10 lists from OmniEmbed (7B), OmniEmbed-mv, and OmniEmbed-mv with RankVideo reranking. Appendix \ref{append:generation} has results in the \textsc{Oracle} setting, as well as further \textsc{RAG} results based on top-20 lists.

\paragraph{Results} \autoref{tab:rag-qwen-reference-k10} shows MiRAGE metrics for each generation baseline on \mvraw under the three retrieval systems discussed above. Both TRACE and MARQUIS tend to outperform CAG, with TRACE\textsubscript{lite} consistently achieving the highest \infof scores and MARQUIS\textsubscript{bullet} tending to obtain the best \citationf scores. Comparing results under different retrievers validates the importance of this part of the pipeline, as evidenced by the large (${\sim}$6--8 point) gaps in \infof performance between the best retrieval system (OmniEmbed-mv with reranking) and the worst (OmniEmbed).
Comparing the same systems across splits, we generally find modestly higher scores on \raw than on \mixed for both the information and citation metrics---consistent with the first-stage retrieval results in \S\ref{subsec:baselines-retrieval}---though this is not universally the case.
%Further, consistent with the nDCG results in \S\ref{subsec:baselines-retrieval}, we do not find meaningful differences in performance between the \mixed and \raw subsets \todo{Wil: not quite accurate, fix}.
Finally, all numbers are very low in absolute terms (especially for the citation metrics), confirming the difficulty of the report generation task.

% MARQUIS\textsubscript{baseline} and TRACE\textsubscript{lite} achieving the highest \infof scores, and with TRACE\textsubscript{lite} definitively obtaining the best \citationf results. For MARQUIS, the best-performing variant differs depending on the metric: MARQUIS\textsubscript{baseline} shows best results on \infof but MARQUIS\textsubscript{bullet} achieves better performance on \citationf. [WHY?]

\section{Conclusion}
\label{sec:conclusion}
We have introduced \mvraw, a benchmark for retrieval and generation over raw, event-centric video that spans 130 events and nearly 120,000 videos. To enable rapid system development on both tasks, we also release \microv---a dev set comprising 23 events and close to 1,000 videos. Further, we present baseline results on both tasks. For retrieval, we show that even strong baselines that pair first-stage dense retrieval methods with recent video rerankers show middling performance (\S\ref{subsec:baselines-retrieval}), testifying to \mvraw's difficulty. We show that raw video retrieval is especially challenging when retrieving over a \emph{mixture} of both raw and edited video. On generation (\S\ref{subsec:baselines-generation}), we see similarly underwhelming performance, where models struggle both with effectively recovering key information for the target queries and with providing appropriate citations. Nonetheless, with the recent rapid improvements in multimodal models, we are confident that further advances in retrieval and reasoning over raw video are in store, and we release \mvraw to track this progress.
{
    \small
    \bibliographystyle{ieeenat_fullname}
    \bibliography{main, anthology-1, anthology-2}
}

\appendix
\clearpage
\setcounter{page}{1}
\maketitlesupplementary

\section{Data}
\label{append:data-details}
\subsection{Summary Statistics}
\label{subappend:summary-stats}

\begin{table}
    \small
    \centering
    \setlength{\tabcolsep}{3pt}
    \begin{tabular}{lll}
    \toprule
         & \bf \microv & \bf \mvraw \\
    \midrule
        Events (topics) & 23 & 130 \\
        Queries & 31 & 222 \\
        Videos & 933 & 118,802 \\
        Total duration (hrs) & 34 & 5,351 \\
        Mean duration (mins) & 2.2 & 2.7 \\
    \bottomrule
    \end{tabular}
    \caption{Summary statistics for \microv and \mvraw (\mixed).}
    \label{tab:full-summary-stats}
\end{table}

\autoref{tab:full-summary-stats} contains summary statistics for \microv and \mvraw (\mixed).

\subsection{Raw Video Classification}
\label{subappend:raw-classifier}
\begin{table}[t]
\centering
\begin{tabular}{lccc}
\toprule
    \textbf{System} & \textbf{Accuracy} & \textbf{Raw-F1} & \textbf{Macro-F1} \\
    \midrule
    ZS v1 & 91.3 & 75.3 & 85.0 \\
    ZS v2 & 91.3 & 72.6 & 83.7 \\
    Features & 90.2 & 71.0 & 82.6 \\
    LoRA-Binary & 93.8 & 79.6 & 88.0 \\
    LoRA-Quad & 93.7 & 81.5 & 88.8 \\
    \midrule
    \textbf{Ensemble} & \textbf{95.1} & \textbf{85.7} & \textbf{91.4} \\
\bottomrule
\end{tabular}
\caption{Raw video classification performance of the individual ensemble members and the final ensemble.}
\label{tab:rawvid}
\end{table}

% \alex{we want to filter the data for some reason (the focus is raw video, we want to see performance of retrieval and generation in RAG when there are only raw videos).}
As noted in \S\ref{subsec:postprocessing}, we train a binary classifier to distinguish raw from edited video and apply this classifier to the \mixed split of \mvraw to obtain the \raw split, which removes all irrelevant videos labeled as \emph{edited}. Here, we provide more details on the development of this classifier.

We use the test split of \textsc{MultiVENT 2.0} as training data for this classifier, which contains labels for four different video types:\footnote{We use the test split because it is the only split that contains these labels.} (1) \emph{raw}: the same as our notion of \emph{purely} raw video; (2) \emph{diet raw}: videos that are nearly fully raw, but that may have minor edited elements (e.g., brief music, a scene cut or two); (3) \emph{edited}: videos that have been substantially edited by an individual; and (4) \emph{professional}: videos that have been substantially edited by a professional news or governmental organization. While one component of our final classifier relies on this four-way categorization, we also construct a binary categorization by grouping together the \emph{raw} and \emph{diet raw} labels (on the one hand) and the \emph{edited} and \emph{professional} labels (on the other). We construct random 85\%/15\% train/dev splits from the original \textsc{MultiVENT 2.0} test split. We use these splits to iterate on classifier design using an autoresearcher tool powered by Claude Opus 4.7 and tasked with optimizing (binary) Macro-F1 on the dev split. The best-performing system we found was a weighted ensemble of the five individual classifiers described below. \autoref{tab:rawvid} shows the ensemble's performance on the dev set.

\begin{enumerate}
    \item \textbf{Features.} A gradient-boosted classifier over statistics extracted with FFmpeg (duration, geometry, frame rate, bitrate and detected shot boundaries), and a pass over the low-resolution grayscale frames that measures overlays, camera motion, and letterboxing and fade transitions. 
    \item \textbf{ZS v1.} Qwen3.5-27B \cite{qwen3.5} zero-shot prompted with the prompt in \autoref{fig:prompt-zsv1}.
    \item \textbf{ZS v2.} Qwen3.5-27B \cite{qwen3.5} zero-shot prompted with the prompt in \autoref{fig:prompt-zsv2}.
    \item \textbf{LoRA-Binary.} A LoRA \cite{hu2022lora} fine-tuned version of Qwen3.5-9B trained on the binary classification scheme (\emph{edited} vs.\ \emph{raw} labels). We fine-tune for one epoch and use the prompt in \autoref{fig:prompt-binary}.
    \item \textbf{LoRA-Quad.} A LoRA \cite{hu2022lora} fine-tuned version of Qwen3.5-9B \cite{qwen3.5} trained on the four-way classification scheme (\emph{raw}, \emph{diet raw}, \emph{edited}, and \emph{professional} labels). The label is predicted by summing the log-likelihood of the \emph{raw} and \emph{diet raw} labels. We fine-tune for one epoch and use the prompt in \autoref{fig:prompt-quad}. 
\end{enumerate}

The weights on the individual classifiers in the ensemble are as follows, with the \emph{raw} score threshold set at $-0.657$:
\begin{equation*}
\begin{split}
\textsc{Raw} = \;& 0.5 \cdot \text{LoRA-Binary} \\
             +\;& 0.125 \cdot \text{LoRA-Quad} \\
             +\;& 0.125 \cdot \text{ZS v1} \\
                 +\;& 0.125 \cdot \text{ZS v2} \\
             +\;& 0.125 \cdot \text{Features}
\end{split}
\end{equation*}

\section{Evaluation}
\label{append:eval-details}
% \todo{Here we provide details about our implementation of the MiRAGE metric suite for generation evaluation \cite{martin2026seeingmirageevaluatingmultimodal}.}

We evaluate the MiRAGE metrics under the ``reference'' (as opposed to ``collection'') setting for all reported numbers. This means we use the reference set of query claims for the target query to assess the support of claims in a predicted report, rather than assessing support directly against the video content. We refer the reader to \cite{martin2026seeingmirageevaluatingmultimodal} for additional discussion of the \emph{reference} vs.\ \emph{collection} distinction. We detail several additional metric implementation decisions below.

\paragraph{Claim Decomposition}
We decompose claims with the prompt in \autoref{fig:decontext-prompt}, adapting methods from \cite{gunjal-durrett-2024-molecular} and \cite{wanner-etal-2025-dndscore} for claim decomposition and decontextualization. This is particularly important for handling long predicted reports, as it enables individual claims to be understood and evaluated independently, without the need to provide long passages as contextualizing information.

\paragraph{Long-Context Predictions}
One of our baseline systems, TRACE, produces exceptionally long predicted reports (often $\geq$10,000 sentences per report). The length of these reports causes issues with context rot and instruction following in our claim verification for the MiRAGE precision and recall metrics. For precision, claim decontextualization helps with this issue, as verifying a predicted claim against the reference claims can be done without the predicted text to help contextualize the claim. For recall, we reformulate \infor to operate over windows of 100 claims. 
\begin{equation*}
    \text{\infor} = \frac{1}{|C_r|}\sum_{c \in C_r} \max_{w \in W(C_p)} \mathbbm{1}\!\left[\textrm{supported}(c \mid w)\right]
\end{equation*}
where $C_r$ is the list of reference claims, $C_p$ is the list of claims decomposed and decontextualized from the predicted report, and $W(C_p)$ is the set of (contiguous) 100-claim windows that (together) cover $C_p$.

\paragraph{Prompts}
We otherwise follow the same MiRAGE implementation as outlined in \cite{martin2026seeingmirageevaluatingmultimodal}, using the prompts in \autoref{fig:info-verification-prompt} and \autoref{fig:cite-verification-prompt} for claim verification.

\section{Generation}
\label{append:generation}
% \subsection{Additional Results}
% \todo{here we will put the reference results of Qwen3.5 and the CLUE results. discuss why some methods have a lower reference precision --> they include a lot of content irrelevant to the queries that \textbf{are} grounded in the videos.}

% \input{tables/multivent-clue-collection-oracle}
% \begin{table}[t]
% \centering
% \small
% \begin{tabular}{lcccccc}
% \toprule
% & \multicolumn{3}{c}{InfoF1} & \multicolumn{3}{c}{CiteF1} \\
% \cmidrule(lr){2-4}\cmidrule(lr){5-7}
% System & P & R & F1 & P & R & F1 \\
% \midrule
% \multicolumn{7}{l}{\emph{MicroVent} ($n=32$ queries)} \\
% CAG & .184 & .262 & .173 & .140 & .179 & .119 \\
% MARQUIS (baseline) & .351 & .238 & .241 & .209 & .144 & .132 \\
% MARQUIS (bullet) & .389 & .231 & .232 & .295 & .216 & .193 \\
% MARQUIS (ginger) & .373 & .225 & .218 & .176 & .125 & .098 \\
% TRACE (ASR-only) & .329 & .446 & .350 & .216 & .351 & .232 \\
% TRACE (full) & .313 & .442 & .339 & .213 & .321 & .225 \\
% \midrule
% \multicolumn{7}{l}{\emph{MultiVENT-mixed} ($n=222$ queries)} \\
% CAG & .156 & .338 & .182 & .124 & .242 & .134 \\
% MARQUIS (baseline) & .295 & .343 & .268 & .170 & .238 & .160 \\
% MARQUIS (bullet) & .310 & .306 & .248 & .221 & .273 & .196 \\
% MARQUIS (ginger) & .292 & .300 & .244 & .141 & .187 & .121 \\
% TRACE (ASR-only) & .260 & .437 & .297 & .179 & .348 & .208 \\
% TRACE (full) & .250 & .379 & .273 & .173 & .291 & .191 \\
% \bottomrule
% \end{tabular}
% \caption{MiRAGE scores with the \textbf{Qwen3.5-27B} judge, \textbf{reference} setting.}
% \label{tab:multivent-qwen-reference}
% \end{table}

\begin{table*}[t]
\centering
\small
\setlength{\tabcolsep}{4pt}
\begin{tabular}{lrrrrrrrrrrrr}
\toprule
& \multicolumn{6}{c}{\bf \microv} & \multicolumn{6}{c}{\bf \mvraw (\raw)} \\
\cmidrule(lr){2-7}\cmidrule(lr){8-13}
& \multicolumn{3}{c}{\bf Information} & \multicolumn{3}{c}{\bf Citation} & \multicolumn{3}{c}{\bf Information} & \multicolumn{3}{c}{\bf Citation} \\
\cmidrule(lr){2-4}\cmidrule(lr){5-7}\cmidrule(lr){8-10}\cmidrule(lr){11-13}
\bf System & \bf P & \bf R & \bf F1 & \bf P & \bf R & \bf F1 & \bf P & \bf R & \bf F1 & \bf P & \bf R & \bf F1 \\
\midrule
CAG                              & 18.4 & 26.2 & 17.3 & 14.0 & 17.9 & 11.9 & 15.6 & 33.8 & 18.2 & 12.4 & 24.2 & 13.4 \\
MARQUIS\textsubscript{CAG}  & 35.1 & 23.8 & 24.1 & 20.9 & 14.4 & 13.2 & 29.5 & 34.3 & 26.8 & 17.0 & 23.8 & 16.0 \\
MARQUIS\textsubscript{bullet}    & 38.9 & 23.1 & 23.2 & 29.5 & 21.6 & 19.3 & 31.0 & 30.6 & 24.8 & 22.1 & 27.3 & 19.6 \\
MARQUIS\textsubscript{ginger}    & 37.3 & 22.5 & 21.8 & 17.6 & 12.5 & 9.8  & 29.2 & 30.0 & 24.4 & 14.1 & 18.7 & 12.1 \\
TRACE\textsubscript{lite}    & 32.9 & 44.6 & \textbf{35.0} & 21.6 & 35.1 & \textbf{23.2} & 26.0 & 43.7 & \textbf{29.7} & 17.9 & 34.8 & \textbf{20.8} \\
TRACE\textsubscript{full}        & 31.3 & 44.2 & 33.9 & 21.3 & 32.1 & 22.5 & 25.0 & 37.9 & 27.3 & 17.3 & 29.1 & 19.1 \\
\bottomrule
\end{tabular}
\caption{MiRAGE metric results (``reference'' variant, using Qwen3.5-27B as judge) on \microv and the \raw split of \mvraw in the \textsc{Oracle} generation setting. The best F1 in each column is bolded.}
\label{tab:multivent-qwen-reference}
\end{table*}

% requires \usepackage{multirow}
\begin{table*}[t]
\centering
\small
\setlength{\tabcolsep}{4pt}
\begin{tabular}{llrrrrrrrrrrrr}
\toprule
& & \multicolumn{6}{c}{\bf\mixed} & \multicolumn{6}{c}{\bf\raw} \\
\cmidrule(lr){3-8}\cmidrule(lr){9-14}
& & \multicolumn{3}{c}{\bf Information} & \multicolumn{3}{c}{\bf Citation} & \multicolumn{3}{c}{\bf Information} & \multicolumn{3}{c}{\bf Citation} \\
\cmidrule(lr){3-5}\cmidrule(lr){6-8}\cmidrule(lr){9-11}\cmidrule(lr){12-14}
\bf Retriever & \bf System & \bf P & \bf R & \bf F1 & \bf P & \bf R & \bf F1 & \bf P & \bf R & \bf F1 & \bf P & \bf R & \bf F1 \\
\midrule
\multirow{6}{*}{OmniEmbed (7B)}
 & CAG                             & 10.2 & 23.7 & 11.4 & 2.1 & 3.9 & \textbf{2.2} & 9.2 & 25.3 & 11.3 & 3.0 & 6.2 & \textbf{3.4} \\
 & MARQUIS\textsubscript{CAG} & 15.4 & 19.9 & 13.6 & 2.0 & 2.6 & 1.5 & 16.1 & 18.7 & 14.1 & 3.4 & 3.8 & 2.2 \\
 & MARQUIS\textsubscript{bullet}   & 20.5 & 26.2 & 19.0 & 1.6 & 4.8 & 1.8 & 19.9 & 24.3 & 17.6 & 2.9 & 6.5 & 2.8 \\
 & MARQUIS\textsubscript{ginger}   & 18.9 & 18.3 & 13.4 & 1.4 & 1.4 & 0.8 & 21.5 & 18.7 & 14.7 & 2.1 & 2.5 & 1.1 \\
 & TRACE\textsubscript{lite}       & 16.3 & 36.2 & \textbf{20.3} & 0.9 & 6.9 & 1.4 & 15.5 & 34.3 & \textbf{19.1} & 1.7 & 10.0 & 2.6 \\
 & TRACE\textsubscript{full}       & 15.3 & 33.3 & 18.8 & 1.0 & 6.1 & 1.5 & 14.3 & 29.2 & 17.0 & 1.8 & 9.5 & 2.5 \\
\cmidrule(lr){1-14}
\multirow{6}{*}{OmniEmbed-mv (7B)}
 & CAG                             & 10.1 & 32.1 & 12.9 & 3.6 & 12.3 & \textbf{4.3} & 10.7 & 31.5 & 13.3 & 4.5 & 13.5 & 5.2 \\
 & MARQUIS\textsubscript{CAG} & 16.6 & 24.5 & 15.9 & 4.2 & 7.4 & 3.1 & 17.9 & 24.2 & 16.4 & 4.9 & 8.5 & 4.0 \\
 & MARQUIS\textsubscript{bullet}   & 21.9 & 34.1 & 22.7 & 3.3 & 13.5 & 4.0 & 24.3 & 33.0 & 23.6 & 5.4 & 15.4 & \textbf{5.6} \\
 & MARQUIS\textsubscript{ginger}   & 21.1 & 20.4 & 15.6 & 2.4 & 3.7 & 1.3 & 25.0 & 21.1 & 17.1 & 3.4 & 4.6 & 1.8 \\
 & TRACE\textsubscript{lite}       & 19.6 & 45.2 & \textbf{25.0} & 2.5 & 19.0 & 4.0 & 19.6 & 45.2 & \textbf{25.1} & 3.3 & 22.5 & 5.2 \\
 & TRACE\textsubscript{full}       & 18.1 & 40.1 & 22.8 & 2.4 & 16.9 & 3.8 & 18.5 & 39.0 & 22.8 & 3.5 & 19.6 & 5.1 \\
\cmidrule(lr){1-14}
\multirow{6}{*}{\ \ + RankVideo (8B)}
 & CAG                             & 10.1 & 32.4 & 12.9 & 3.4 & 12.2 & 4.2 & 10.0 & 32.4 & 12.6 & 4.1 & 14.1 & 5.0 \\
 & MARQUIS\textsubscript{CAG} & 17.6 & 25.4 & 16.9 & 4.9 & 8.2 & 3.8 & 19.2 & 25.0 & 17.4 & 6.0 & 9.1 & 4.6 \\
 & MARQUIS\textsubscript{bullet}   & 25.7 & 33.8 & 24.2 & 4.6 & 14.7 & \textbf{4.9} & 25.9 & 31.8 & 23.7 & 6.2 & 16.0 & \textbf{6.4} \\
 & MARQUIS\textsubscript{ginger}   & 24.0 & 21.1 & 17.0 & 2.7 & 4.1 & 1.3 & 25.6 & 20.6 & 17.0 & 3.8 & 4.2 & 2.1 \\
 & TRACE\textsubscript{lite}       & 20.8 & 46.5 & \textbf{26.4} & 2.7 & 21.7 & 4.4 & 20.8 & 45.9 & \textbf{26.2} & 3.4 & 23.9 & 5.4 \\
 & TRACE\textsubscript{full}       & 19.0 & 42.7 & 24.1 & 2.8 & 19.0 & 4.3 & 19.8 & 40.2 & 24.3 & 3.6 & 20.6 & 5.3 \\
\bottomrule
\end{tabular}
\caption{MiRAGE metric results (``reference'' variant, using Qwen3.5-27B as judge) on the \mixed and \raw splits of \mvraw, given the top $k=20$ videos from three different retrieval systems. Retriever names follow Table~\ref{tab:mvraw-retrieval-baselines}, which reports the corresponding retrieval effectiveness.
%All conditions share one query set, one reference and one authoritative collection, so cells are directly comparable.
The best F1 in each column within a retriever block is bolded.}
\label{tab:rag-qwen-reference-k20}
\end{table*}

In \autoref{tab:multivent-qwen-reference}, we report performance of the baseline generation systems in the \textsc{Oracle} setting on \microv and the \raw split of \mvraw. In \autoref{tab:rag-qwen-reference-k20}, we report performance of the baseline generation systems using the top $k=20$ videos (instead of $k=10$ reported in \autoref{tab:rag-qwen-reference-k10} in the main text).

\section{Retrieval}
\label{append:retrieval}
\subsection{Gemma 4 Reranking}
Zero-shot reranking with Gemma 4 scores each query--video pair independently.
Gemma's input consists of video frames followed by a reranking prompt.
Videos are represented as 8 uniformly sampled frames with a maximum of 280 visual tokens per frame.
The prompt is shown in \autoref{app:gemma-rerank-prompt}.
Each video's relevance score is the
yes--no log-odds read from the first generated token position:
$\log\!\sum_{t\in\{\text{yes},\text{Yes}\}}\!p(t) -
\log\!\sum_{t\in\{\text{no},\text{No}\}}\!p(t)$.

\begin{figure}[t]
\centering
\promptbox{%
Query: \{query\}

Is this video relevant to the query? Answer \textbf{yes} or \textbf{no}.%
}
\caption{Zero-shot \texttt{gemma-4-12B-it} video reranking prompt.}
\label{app:gemma-rerank-prompt}
\end{figure}

\begin{figure}[t]
\centering
\promptbox{%
You are shown 36 frames sampled evenly across the whole video, tiled into
$3\times3$ grids. Read each grid left-to-right, top-to-bottom; that is
chronological order.

Classify the production style. Be careful: lightly-edited amateur footage
counts as \textbf{Raw}, not \textbf{Edited}.

\begin{promptlist}
  \item \textbf{Raw:} amateur, bystander or single-source footage of an event
        as it happened. One continuous take, or at most a light trim. It may
        carry ONE caption, a watermark, a social-media handle, or a title card
        and still be Raw. Handheld, shaky, vertical, poorly lit, or a fixed
        security/dashcam view are all strong Raw signals.
  \item \textbf{Edited:} a produced piece. Multiple shots cut together, b-roll
        intercut with a presenter, an anchor in a studio, a news graphics
        package (lower-third banners, chyrons, animated titles, station logo),
        or polished broadcast camera work on a tripod.
\end{promptlist}

Ask yourself: would a news editor have had to assemble this from multiple
sources? If no, answer Raw.

Answer with exactly one of: \textbf{Edited}, \textbf{Raw}.%
}
\caption{Prompt for \textbf{ZS v1}.}
\label{fig:prompt-zsv1}
\end{figure}

\begin{figure}[t]
\centering
\promptbox{%
You are shown 36 frames sampled evenly across the whole video, tiled into
$3\times3$ grids. Read each grid left-to-right, top-to-bottom; that is
chronological order. A change of scene between adjacent cells is a cut.

You are classifying the production style of a video. Watch the video and
decide which one of these two categories it belongs to:

\begin{promptlist}
  \item \textbf{Edited:} Post-produced or professionally captured footage:
        cuts, graphics, captions, or a voiceover, or broadcast-quality camera
        work (the Edited and Professional categories).
  \item \textbf{Raw:} Unedited or barely-edited amateur or bystander footage:
        a single continuous take, at most lightly trimmed or carrying one
        caption/watermark (the Raw and Diet Raw categories).
\end{promptlist}

Answer with exactly one of: \textbf{Edited}, \textbf{Raw}.%
}
\caption{Prompt for \textbf{ZS v2}.}
\label{fig:prompt-zsv2}
\end{figure}

\begin{figure}[t]
\centering
\promptbox{%
You are classifying the production style of a video. Watch the video and
decide which one of these two categories it belongs to:

\begin{promptlist}
  \item \textbf{Edited:} Post-produced or professionally captured footage:
        cuts, graphics, captions, or a voiceover, or broadcast-quality camera
        work (the Edited and Professional categories).
  \item \textbf{Raw:} Unedited or barely-edited amateur or bystander footage:
        a single continuous take, at most lightly trimmed or carrying one
        caption/watermark (the Raw and Diet Raw categories).
\end{promptlist}

Answer with exactly one of: \textbf{Edited}, \textbf{Raw}.%
}
\caption{Prompt for \textbf{LoRA-Binary}. The model generates one of the two label strings, which are scored by
log-likelihood.}
\label{fig:prompt-binary}
\end{figure}

\begin{figure}[t]
\centering
\promptbox{%
You are classifying the production style of a video. Watch the video and
decide which one of these four categories it belongs to:

\begin{promptlist}
  \item \textbf{Edited:} Post-produced footage that stitches multiple shots
        together with cuts, graphics, captions, lower-thirds, or a voiceover
        (e.g.\ a news package or montage).
  \item \textbf{Professional:} Professionally captured, broadcast-quality
        footage from a single source (steady framing, good lighting/audio)
        with little or no post-production.
  \item \textbf{Raw:} Unedited amateur or bystander footage: a single
        continuous take from a phone or fixed camera, no cuts, graphics, or
        captions.
  \item \textbf{Diet Raw:} Essentially raw footage with only minimal
        editing\,\textemdash\,light trimming, a single caption/watermark, or
        a title card\,\textemdash\,but no real post-production.
\end{promptlist}

Answer with exactly one of: \textbf{Edited}, \textbf{Professional},
\textbf{Raw}, \textbf{Diet Raw}.%
}
\caption{Prompt for \textbf{LoRA-Quad}. Binary (\emph{raw} vs.\ \emph{edited}) labels are predicted by summing the log-likelihoods of the \emph{raw} and \emph{diet raw} labels.}
\label{fig:prompt-quad}
\end{figure}

\begin{figure*}[p]
\centering
\begin{promptenv}
Instructions:
\begin{promptlist}
  \item You are given a sentence from a longer passage, along with the full
        passage for context.
  \item Your task is to decompose the sentence into atomic claims AND
        decontextualize each claim.
\end{promptlist}

Decomposition rules:
\begin{promptlist}
  \item If the sentence IS atomic (a single, indivisible fact), keep it as-is.
  \item If it's NOT atomic, split into atomic subclaims.
  \item Each subclaim must contain only information present in the original
        sentence.
  \item When in doubt, do NOT decompose. Prefer keeping the claim intact over
        over-splitting.
\end{promptlist}

Decontextualization rules:
\begin{promptlist}
  \item Resolve ALL pronouns (he, she, they, it, etc.) to their referents using
        the full passage.
  \item Resolve demonstratives (this, that, these, those) and other anaphoric
        references.
  \item Replace abbreviated or implicit references with their full form from
        the passage.
  \item Each decontextualized claim must be completely understandable WITHOUT
        the passage.
  \item If the claim is already self-contained, the decontextualized version
        should be identical.
\end{promptlist}

Output ONLY a JSON list with both the original atomic claim and its
decontextualized version:\newline
[\{"claim": "<original atomic claim>", "decontextualized\_claim":
"<self-contained version>"\}, ...]

\textbf{-{}-{}- Examples -{}-{}-}

Full passage for context: Alex had been running for hours through the forest.
He was exhausted and could barely stand. The tall trees around him provided
some shade.

\#\#CLAIM\#\#: He was exhausted and could barely stand.\newline
\#\#DECOMPOSITION\#\#:\newline
\jsonfence json\newline
[\newline
\hspace*{2em}\{"claim": "He was exhausted.", "decontextualized\_claim": "Alex
was exhausted."\},\newline
\hspace*{2em}\{"claim": "He could barely stand.", "decontextualized\_claim":
"Alex could barely stand."\}\newline
]\newline
\jsonfence

\#\#CLAIM\#\#: The tall trees around him provided some shade.\newline
\#\#DECOMPOSITION\#\#:\newline
\jsonfence json\newline
[\{"claim": "The tall trees around him provided some shade.",
"decontextualized\_claim": "The tall trees around Alex in the forest provided
some shade."\}]\newline
\jsonfence

Full passage for context: On April 15, 2019, Notre-Dame de Paris caught fire.
The blaze destroyed the spire and most of the roof. It took over 400
firefighters to contain the flames.

\#\#CLAIM\#\#: It took over 400 firefighters to contain the flames.\newline
\#\#DECOMPOSITION\#\#:\newline
\jsonfence json\newline
[\{"claim": "It took over 400 firefighters to contain the flames.",
"decontextualized\_claim": "It took over 400 firefighters to contain the
flames of the Notre-Dame de Paris fire."\}]\newline
\jsonfence

Full passage for context:\newline
[context]

\#\#CLAIM\#\#: [claim]\newline
\#\#DECOMPOSITION\#\#:
\end{promptenv}
\caption{Prompt used to decompose and decontextualize claims.}
\label{fig:decontext-prompt}
\end{figure*}

\begin{figure*}[t]
\centering
\promptlabel{System}
\begin{promptenv}
You are an expert in evaluating and verifying claims. You will be given a
passage of text and a claim. Your task is to determine if the claim is
supported by the passage of text. Output exactly one lowercase token:
\textbf{yes} if the claim is supported by the passage, or \textbf{no} if the
claim is not supported by the passage.
\end{promptenv}
\promptlabel{User}
\begin{promptenv}
Here is the passage: <verification\_context> [PUT\_VERIFICATION\_CONTEXT\_HERE]
<verification\_context>.\newline
Here is the claim: <claim> [PUT\_CLAIM\_HERE] <claim>

Output exactly one lowercase token: \textbf{yes} or \textbf{no}. Is the claim:
[PUT\_CLAIM\_HERE], supported by the passage?
\end{promptenv}
\caption{Information verification prompt (\infop and \infor, reference setting).
\texttt{[PUT\_VERIFICATION\_CONTEXT\_HERE]} receives the reference claim set for
\infop, and one $100$-claim window of the predicted report for \infor.}
% Note that no slot exists for the source report: the claim is judged on its own text, which is what makes the metric robust to a submission that pads its report in order to ground its own claims.}
\label{fig:info-verification-prompt}
\end{figure*}

\begin{figure*}[t]
\centering
\promptlabel{System}
\begin{promptenv}
You are an expert in evaluating and verifying claims. You will be given a
claim and a list of claims to verify the claim against. Your task is to
determine if the claim is supported by the list of claims. Output exactly one
lowercase token: \textbf{yes} if the claim is supported by the list of claims,
or \textbf{no} if the claim is not supported by the list of claims.
\end{promptenv}
\promptlabel{User}
\begin{promptenv}
Here is the list of claims to verfiy against: <verification\_context>
[PUT\_VERIFICATION\_CONTEXT\_HERE] <verification\_context>.\newline
Here is the claim: <claim> [PUT\_CLAIM\_HERE] <claim>

Output exactly one lowercase token: \textbf{yes} or \textbf{no}. Is the claim:
[PUT\_CLAIM\_HERE], supported by list of claims to verify against?
\end{promptenv}
\caption{Citation verification prompt (\citationp and \citationr, reference setting). Here, the verification context is given as a list of claims. This prompt is
issued once per cited source and the results are aggregated as a logical OR.}
\label{fig:cite-verification-prompt}
\end{figure*}

\end{document}